\documentclass{article}
\usepackage{verbatim,spconf,amsmath,graphicx,amssymb,xcolor}

\title{Pixel-level Corrosion Detection on Metal Constructions by Fusion of Deep Learning Semantic and Contour Segmentation}

\name{Iason Katsamenis\textsuperscript{1}, Eftychios Protopapadakis\textsuperscript{1}, Anastasios Doulamis\textsuperscript{1},}
\secondlinename{Nikolaos Doulamis\textsuperscript{1} and Athanasios Voulodimos\textsuperscript{2}}

\address{\textsuperscript{1} National Technical University of Athens, 9\textsuperscript{th}, Heroon Polytechniou str. 15780 Athens, Greece \\\textsuperscript{2} University of West Attica, 28\textsuperscript{th}, Agiou Spyridonos str. 12243 Athens, Greece\thanks{This paper is supported by the H2020 PANOPTIS project “Development of a Decision Support System for Increasing the Resilience of Transportation Infrastructure based on combined use of terrestrial and airborne sensors and advanced modelling tools,” under grant agreement 769129.}}

\begin{document}
\maketitle
\begin{abstract}
Corrosion detection on metal constructions is a major  challenge in civil engineering for quick, safe and effective inspection. Existing image analysis approaches tend to place bounding boxes around the defected region which is not adequate both for structural analysis and pre-fabrication, an innovative construction concept which reduces maintenance cost, time and improves safety. In this paper, we apply three semantic segmentation-oriented deep learning models (FCN, U-Net and Mask R-CNN) for  corrosion detection, which perform better in terms of accuracy and time and require a smaller number of annotated samples compared to other deep models, e.g. CNN. However, the final images derived are still not sufficiently accurate for structural analysis and pre-fabrication. Thus, we adopt a novel data projection scheme that fuses the results of color segmentation, yielding accurate but over-segmented contours of a region, with a processed area of the deep masks, resulting in high-confidence corroded pixels. 
\end{abstract}
\section{Introduction}
\label{sec:intro}

Metal constructions are widely used in transportation infrastructures, including bridges, highways and tunnels. Rust and corrosion may result in severe problems in safety. Hence, metal defect detection is a major challenge in civil engineering to achieve quick, effective but also safe inspection, assessment and maintenance of the infrastructure \cite{liu2017image} and deal with materials’ deterioration phenomena that derive from several factors, such as climate change, weather events and ageing. 

Current approaches in image analysis for detecting defects are through bounding boxes placed around defected areas to assist engineers to rapidly focus on damages \cite{protopapadakis2019automatic,Protopapadakis2017IISA,cha2017deep,soukup2014convolutional}. 
Such approaches, however, are not adequate for a structural analysis since several metrics (e.g. area, aspect ratio, maximum distance) are required to assess the defect status. Thus, we need a more precise pixel-level classification which can also trigger the novel ideas in construction of {\it pre-fabrication} \cite{dong2017study}.  
Pre-fabrication allows components to be built outside the infrastructure, decreasing maintenance cost and time, and improving traffic flows and working risks. Additionally, real-time classification response is necessary to achieve fast inspection of the critical infrastructure, especially on large-scale structures. Finally, a small number of training samples is available, due to the fact that specific traffic arrangements, specialized equipment and extra manpower are required, increasing the cost dramatically.

\subsection {Related Work}

Currently, deep learning algorithms \cite{voulodimos2018deep} have been proposed for defect detection. Since the data received as 2D image inputs, convolutional neural networks (CNNs) have been applied to identify regions of interest \cite{protopapadakis2019automatic}. Other approaches exploit the CNN structure to detect cracks in concrete and steel infrastructures \cite{doulamis2018combined,cha2017deep}, road damages \cite{protopapadakis2020multi,zhang2016road} and railroad defects on metal surfaces \cite{soukup2014convolutional}. Finally, the work of \cite{chen2017nb} combines a  CNN and a Na\"{i}ve Bayes data fusion scheme to detect crack patches on nuclear power plants.

\begin{figure*}[!htb]
  \includegraphics[width=17cm]{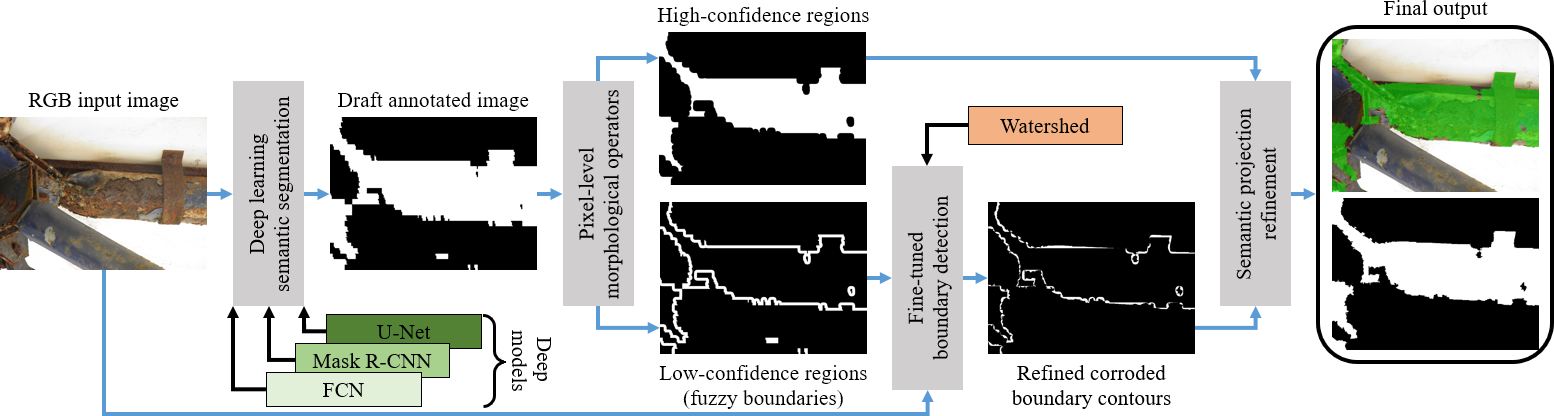}
  \centering
  \caption{An overview of the proposed methodology flowchart.}
  \label{fig:flowchart}
\end{figure*}

The main problem of all the above-mentioned approaches is that they employ conventional deep models, such as CNNs, which require a large number of annotated data  \cite{makantasis2018tensor,Makantasis2018ICASSP}. In our case, such a collection is an arduous task since the annotation should be carried out at pixel level by experts. For this reason, most of existing methods estimate the defected regions through boundary boxes. In addition, the computational complexity of the above methods is high, a crucial factor when inspecting large-scale infrastructures.  

To address these constraints, we exploit alternative approaches in deep learning proposed for semantic segmentation but for applications different than defect detection in transportation networks such as Fully Convolutional Networks (FCN) \cite{long2015fully}, U-Nets \cite{ronneberger2015u} and Mask R-CNN \cite{he2017mask}. The efficiency of the specific methods has been already verified in medical imaging (e.g. brain tumour and COVID-19 symptoms detection) \cite{dong2017automatic,vuola2019mask,voulodimos2020deep}.

\subsection {Paper Contribution}
In this paper, we apply three semantic segmentation-oriented deep models (FCN, U-Net and Mask R-CNN) to detect corrosion in metal structures since they perform more efficiently than traditional deep models. However, the masks derived are still inadequate for structural analysis and pre-fabrication because salient parts of a defected region, especially at the contours, are misclassified. Thus, a detailed, pixel-based mask should be extracted so that civil engineers can take precise measurements on it. 

To overcome this problem, we combine, through projection, the results of color segmentation, which yields accurate contours but oversegments regions, with a processed area of the deep masks (through morphological operators), which indicate only high confident pixels of a defected area. The projection merges color segments belonging to a damaged area improving pixel-based classification accuracy. Experimental results on real-life corroded images, captured in European H2020 Panoptis project, prove the outperformance of the proposed scheme than using segmentation-oriented deep networks or traditional deep models.

\section{The Proposed Overall Architecture}
\label{sec:format}

Let $I \in \mathbb{R}^{w \times h \times 3}$ an RGB image of size $w \times h$. 
Our problem involves a traditional binary classification: areas with intense corrosion grades (rust grade categories B, C and D) and areas of no or minor corrosion (category A). The rust grade categories stems from the standard ISO 8501-1 of civil engineering and are described in Section \ref{sec:datadescr} and depicted in Fig. \ref{fig:rustgrades}. 

Fig. \ref{fig:flowchart} depicts the overall architecture of our approach. The RGB images are fed as inputs to the FCN, U-Net and Mask R-CNN deep models to carry out the semantic segmentation. Despite the effectiveness of these networks, inaccuracies still appear on the contours of the detected objects. Although these errors are small, when one measures them as a percentage of the total corroded region, they are very important for structural analysis and pre-fabrication.

To increase pixel-level accuracy of the derived masks, we combine color segmentation with the regions of the deep models. Color segmentation precisely localizes the contours of an object, but it over-segments it into multiple color areas. Instead, the masks of the deep networks correctly localize the defects, but fail to accurately segment the boundaries. Therefore, we shrink the masks of the deep models to find out the most confident regions, i.e., pixels indicating a defect with high probability. This is done through an erosion morphological operator applied on the initial detections. We also morphologically dilate the deep model regions to localize vague areas  which we need to decide in what region they belong to. On that {\it extended mask}, we apply the watershed segmentation to generate color segments. Finally, we project the results of the color segmentation onto the high confident regions to merge together different color clusters of the same corrosion. 

\section{Deep Semantic Segmentation Models}
\label{sec:SSUDL}
Three types of deep networks are applied to obtain the semantic segments. The first is a Fully Convolutional Network (FCN) \cite{long2015fully} which does not have any fully-connected layers, reducing the loss of spatial information and allowing faster computation. The second is a U-Net built for medical imaging segmentation \cite{ronneberger2015u}.
The architecture is heavily based on FCN, though they have some key differences: U-Net (i) is symmetrical by having multiple upsampling layers and (ii) uses skip connections that apply a concatenation operator instead of adding up. Finally, the third model is the Mask R-CNN \cite{he2017mask}, which extends the Faster R-CNN by using a FCN. This model is able to define bounding boxes around the corroded areas and then segments the rust inside the predicted boxes.

To detect the defects, the models receive as input RGB data and generate, as outputs, binary masks, providing a pixel-level corrosion detection. However, the models fail to generate high fidelity annotations on a boundary level; contours over the detected regions fail to fully encapsulate the rusted regions of the object. As such, a region-growing approach, over these low confidence boundary regions, is applied to improve outcome's robustness and provide refined masks. For training the models, we use an annotated dataset which have been built by civil engineers under the framework of EU project H2020 PANOPTIS.

\section{Refined Detection by Projection - Fusion with a Color Segmentation}
\label{sec:postprocessing}
The presence of inaccuracies in the contours of outputs of the aforementioned deep models is due to the multiple down/up-scaling processes within the convolutional layers of models. To refine the initially detected masks, the following steps are adopted: (i) Localizing a region of high-confident pixels to belong to a defect as a subset of the deep masked outputs. (ii) Localizing fuzzy regions which we cannot decide with confidence if they belong to a corroded area or not, through an extension of the deep output masks. (iii) Applying a color segmentation algorithm in the extended masks which contains both the fuzzy and the high-confident regions. (iv) Finally projecting the results of the color segmentation onto the high-confident area. The projection retains the accuracy in the contours (stemming from color segmentation) while merging different color segments of the same defect together.

\begin{figure}[htb]
\begin{minipage}[b]{.24\linewidth}
  \centering
  \centerline{\includegraphics[width=2.1cm]{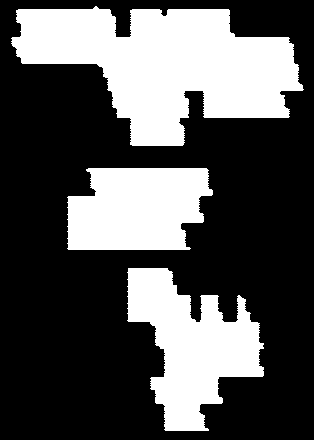}}
  \centerline{(a)}\medskip
\end{minipage}
\hfill
\begin{minipage}[b]{0.24\linewidth}
  \centering
  \centerline{\includegraphics[width=2.1cm]{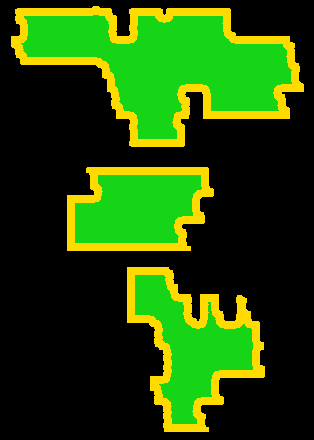}}
  \centerline{(b)}\medskip
\end{minipage}
\hfill
\begin{minipage}[b]{0.48\linewidth}
  \centering
  \centerline{\includegraphics[width=4.2cm]{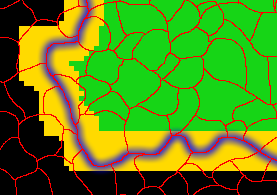}}
  \centerline{(c)}\medskip
\end{minipage}
\caption{Illustrating proposed projection method. (a) Detected regions of interest $R_j$ (white). (b),(c) High confidence $R_j^T$ (green) and fuzzy regions $R_j^{F+}$ (yellow). (c) Color segments $R_j^{w}$ (red) and the final refined contour $R_j^{(a)}$ (blue).}
\label{fig:sure_fuzzy_regions}
\end{figure}

Let us assume that, in an RGB image $I \in \mathbb{R}^{w \times h \times 3}$, we have a set of $N$ corroded regions $R=\{R_1,...,R_N\}$, generated using a deep learning approach. Each partition is a set of pixels $R_j=\{(x_i,y_i)\}_{i=1}^{m_j}$, ${j=1,...,N}$, where $m_j$ is the number of pixels in each set. The remaining pixels represent the no-detection or background areas, denoted by $R^B$, so that $R \cup R^B=I$ (see Fig.~\ref{fig:sure_fuzzy_regions}a). Subsequently, for each $R_j$, ${j=1,...,N}$, we consider two subsets $R_j^T$ and $R_j^F$, so that $R_j^T \cup R_j^F=R_j$. The first set $R_j^T$, corresponds to inner pixels of the region $R_j$, which are considered as true
foreground and indicate high-confidence corroded areas. They can be obtained using an erosion morphological operator on $R_j$. The second set $R_j^F$, contains the remaining pixels $(x_i,y_i) \in R_j$ and their status is considered fuzzy.

We now define a new region $R_j^{F+}	\supset R_j^{F}$, which is a slightly extended area of pixels of $R_j^{F}$, obtained using the dilation morphological operator. The implementation is carried out so that $R_j^T$ and $R_j^{F+}$ are adjacent, but non-overlapping. That is, $R_j^T \cap R_j^{F+}=\varnothing$. Summarizing, we have three sets of areas (see Fig.~\ref{fig:sure_fuzzy_regions}b): (i) True foreground or corroded areas $R^T=\{R_1^T,...,R_N^T\}$ (green region in Fig. \ref{fig:sure_fuzzy_regions}b), (ii) fuzzy areas $R^{F+}=\{R_1^{F+},...,R_N^{F+}\}$ (yellow region of Fig. \ref{fig:sure_fuzzy_regions}b) and (iii) the remaining image areas $R^B$, denoting the background or no-detection areas (black region of Fig. \ref{fig:sure_fuzzy_regions}b).

In the extended fuzzy region $R^{F+}$, we apply the watershed color segmentation algorithm \cite{beucher1992watershed}. 
Let us assume that the color segmentation algorithm produces $M_j$ segments $s_{i,j}$, with $i=1,...,M_j$ for the $j$-th defected region (see the red region in Fig. \ref{fig:sure_fuzzy_regions}c), all of which form a set $R_j^w=\bigcup\limits_{i=1}^{M_j}s_{i,j}$. Then, we project segments $R_j^w$ onto $R_j^T$ in a way that:
\begin{equation}
\centering
R_j^{(a)}=\{s_{i,j}\in R_j^w: s_{i,j}\cap R_j^T \neq \varnothing \}
\end{equation} 

Ultimately, the final detected region (see the blue region in Fig. \ref{fig:sure_fuzzy_regions}c) is defined as the union of all sets $R_j^{(a)}$ over all corroded regions $j=1,...,N$.

\section{Experimental Evaluation}
\label{sec:experimental}

\subsection{Dataset Description}
\label{sec:datadescr}
The dataset is obtained from heterogeneous sources (DSLR cameras, UAVs, cellphones) and contains 116 images of various resolutions, ranging from 194$\times$259 to 4248$\times$2852.  All data have been collected under the framework of H2020 Panoptis project. For the dataset, 80\% is used for training and validation, while the remaining 20\% is for testing. Among the training data, 75\% of them is used for training and the remaining 25\% for validation. The images vary in terms of corrosion type, illumination conditions (e.g. overexposure, underexposure) and environmental landscapes (e.g. highways, rivers, structures). Furthermore, some images contain various types of occlusions, making detection more difficult. 

\begin{figure}[htb]
\begin{minipage}[b]{.24\linewidth}
  \centering
  \centerline{\includegraphics[width=2cm]{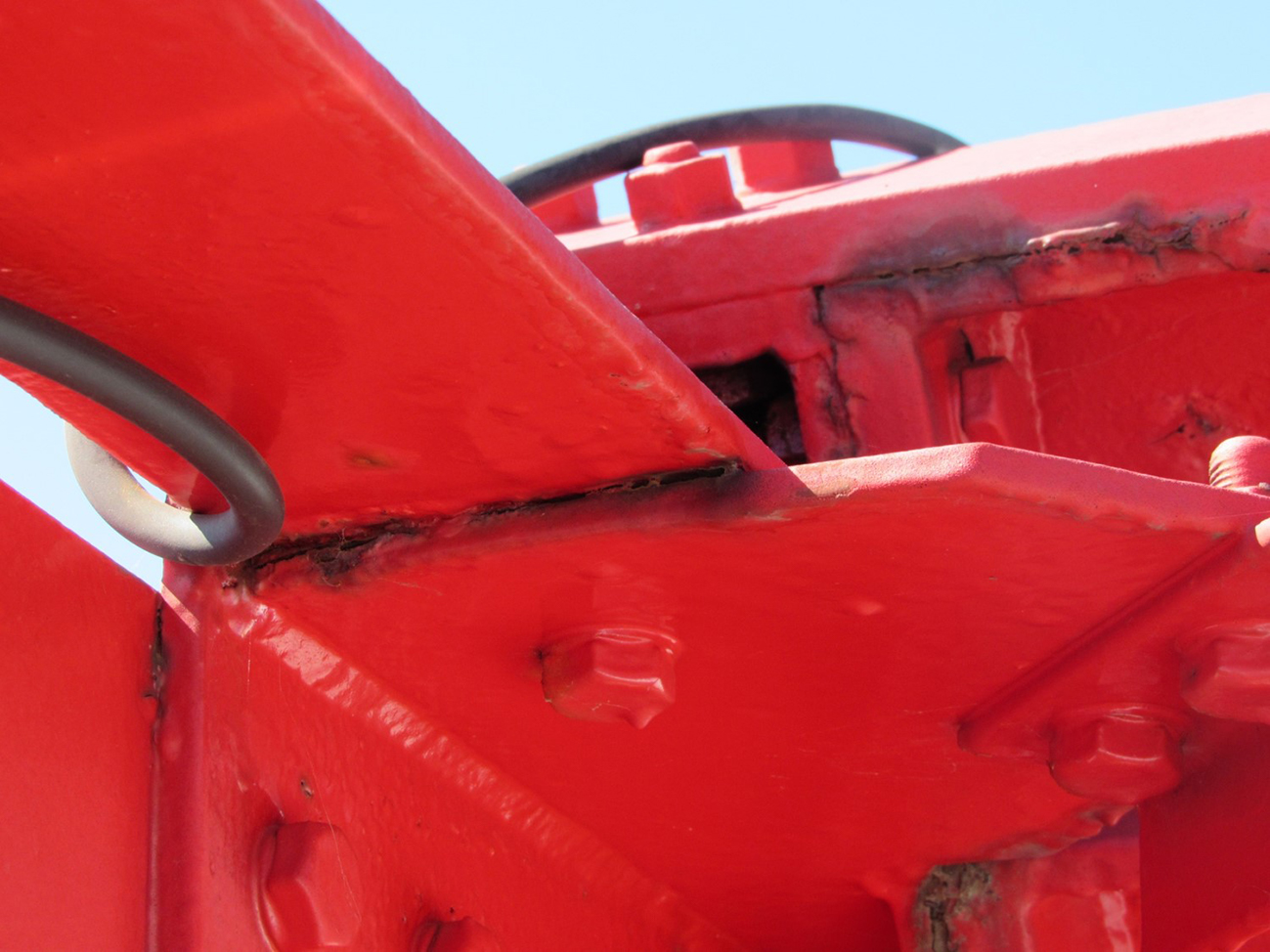}}
  \centerline{(a) Type A}\medskip
\end{minipage}
\hfill
\begin{minipage}[b]{0.24\linewidth}
  \centering
  \centerline{\includegraphics[width=2cm]{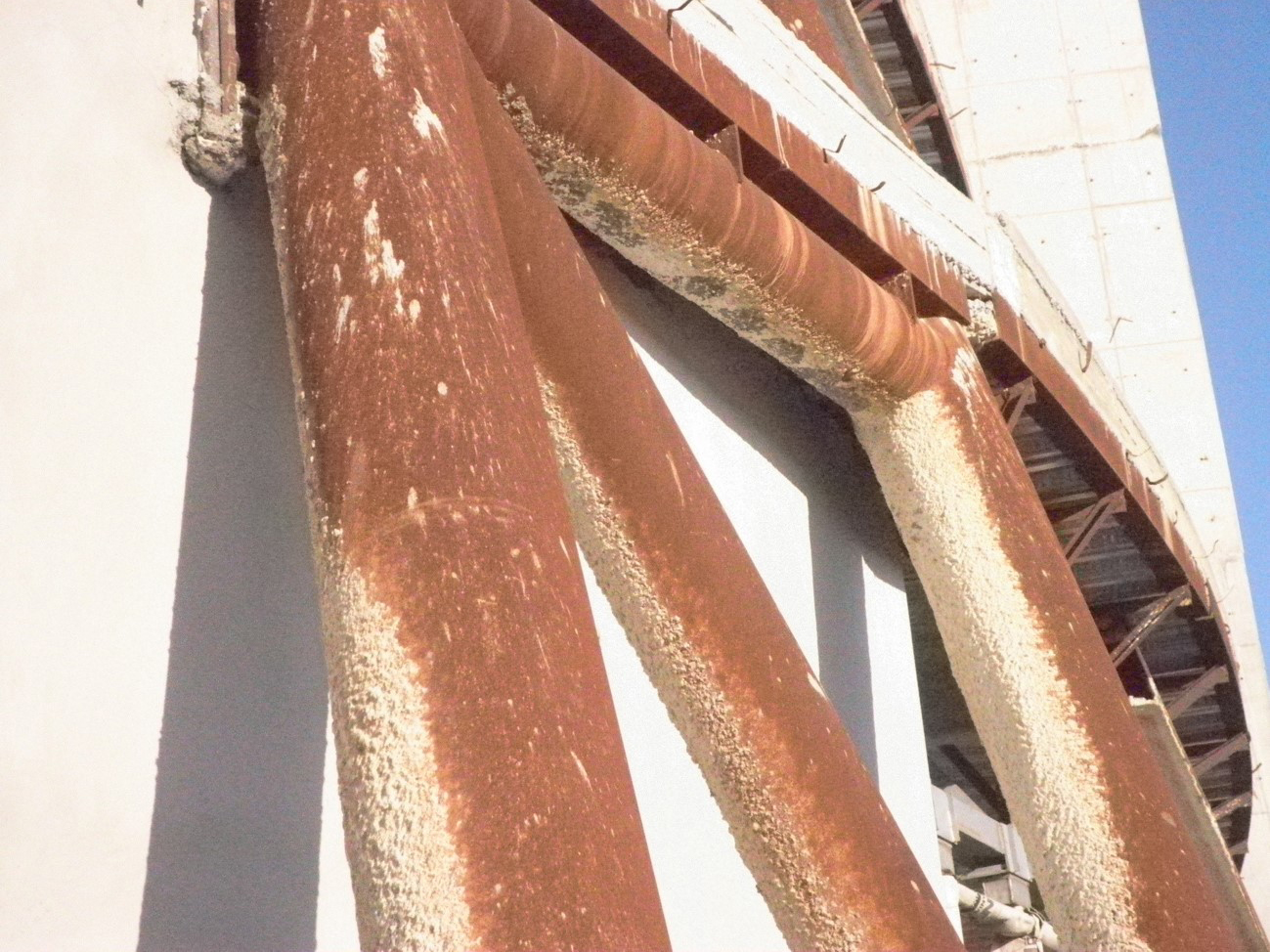}}
  \centerline{(b) Type B}\medskip
\end{minipage}
\hfill
\begin{minipage}[b]{.24\linewidth}
  \centering
  \centerline{\includegraphics[width=2cm]{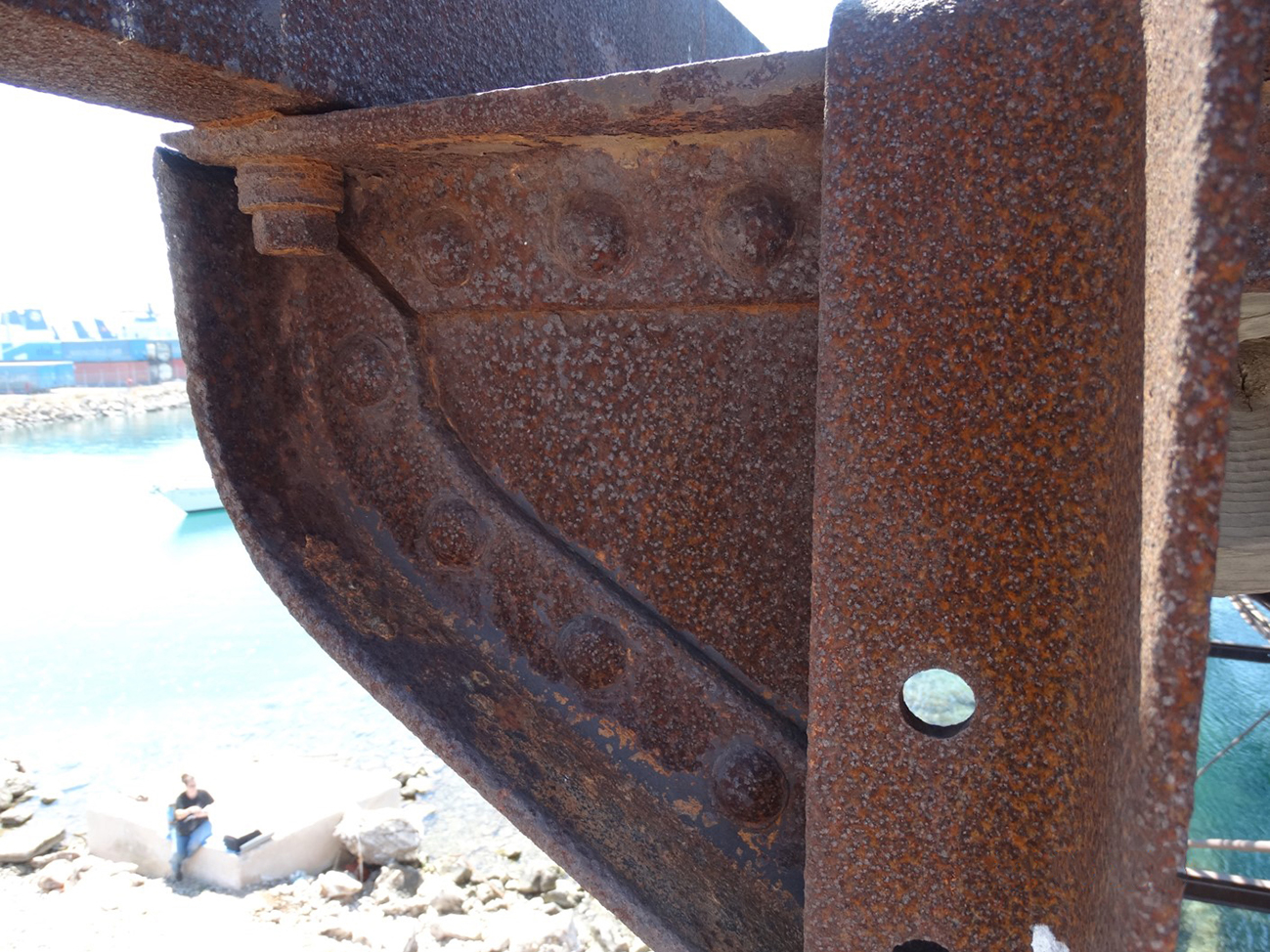}}
  \centerline{(c) Type C}\medskip
\end{minipage}
\hfill
\begin{minipage}[b]{0.24\linewidth}
  \centering
  \centerline{\includegraphics[width=2cm]{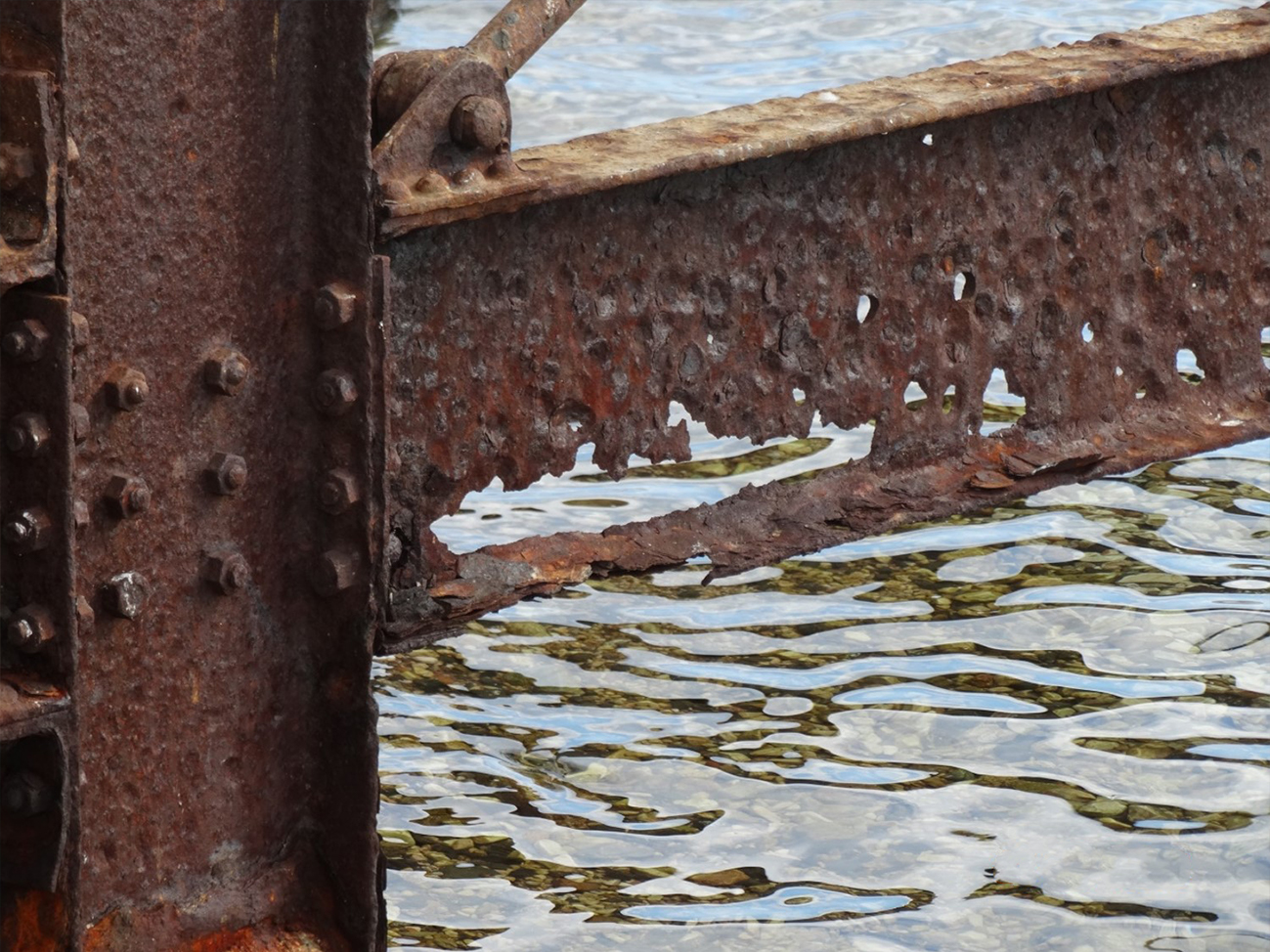}}
  \centerline{(d) Type D}\medskip
\end{minipage}
\caption{Representative image examples of rust grades.}
\label{fig:rustgrades}
\end{figure}

All images of the dataset were manually annotated by engineers within Panoptis project. Particularly, corrosion has been classified  according to the ISO 8501-1 standard (see Fig. \ref{fig:rustgrades}):  
(i) Type A: Steel surface largely covered with adhering mill scale but little, if any, rust.
(ii) Type B: Steel surface which has begun to rust and from which the mill scale has begun to flake.
(iii) Type C: Steel surface on which the mill scale has rusted away or from which it can be scraped, but with slight pitting visible under normal vision.
(iv) Type D: Steel surface on which the mill scale has rusted away and on which general pitting is visible under normal vision.

\subsection{Models Setup}
\label{sec:ModSet}

A common case is the use of pretrained networks of specified topology. Generally, transfer learning techniques serve as starting points, allowing for fast initialization and minimal topological interventions. The FCN-8s variant \cite{long2015fully,shuai2016improving}, served as the main detector for the FCN model. Additionally, the Mask R-CNN detector was based on Inception V2 \cite{szegedy2016rethinking}, pretrained over COCO \cite{lin2014microsoft} dataset. On the other hand, U-Net was designed from scratch. The contracting part, of the adopted variation, had the following setup: Input $\rightarrow$ 2@Conv $\rightarrow$ Pool $\rightarrow$ 2@Conv $\rightarrow$ Pool $\rightarrow$ 2@Conv $\rightarrow$ Drop $\rightarrow$ Pool, where 2@Conv denotes that two consecutive convolutions, of size $3 \times 3$, took place. Finally, for the decoder three corresponding upsampling layers were used.

\begin{figure}[h]
\begin{minipage}[b]{0.18\linewidth}
  \centering
  \centerline{\includegraphics[width=1.6cm]{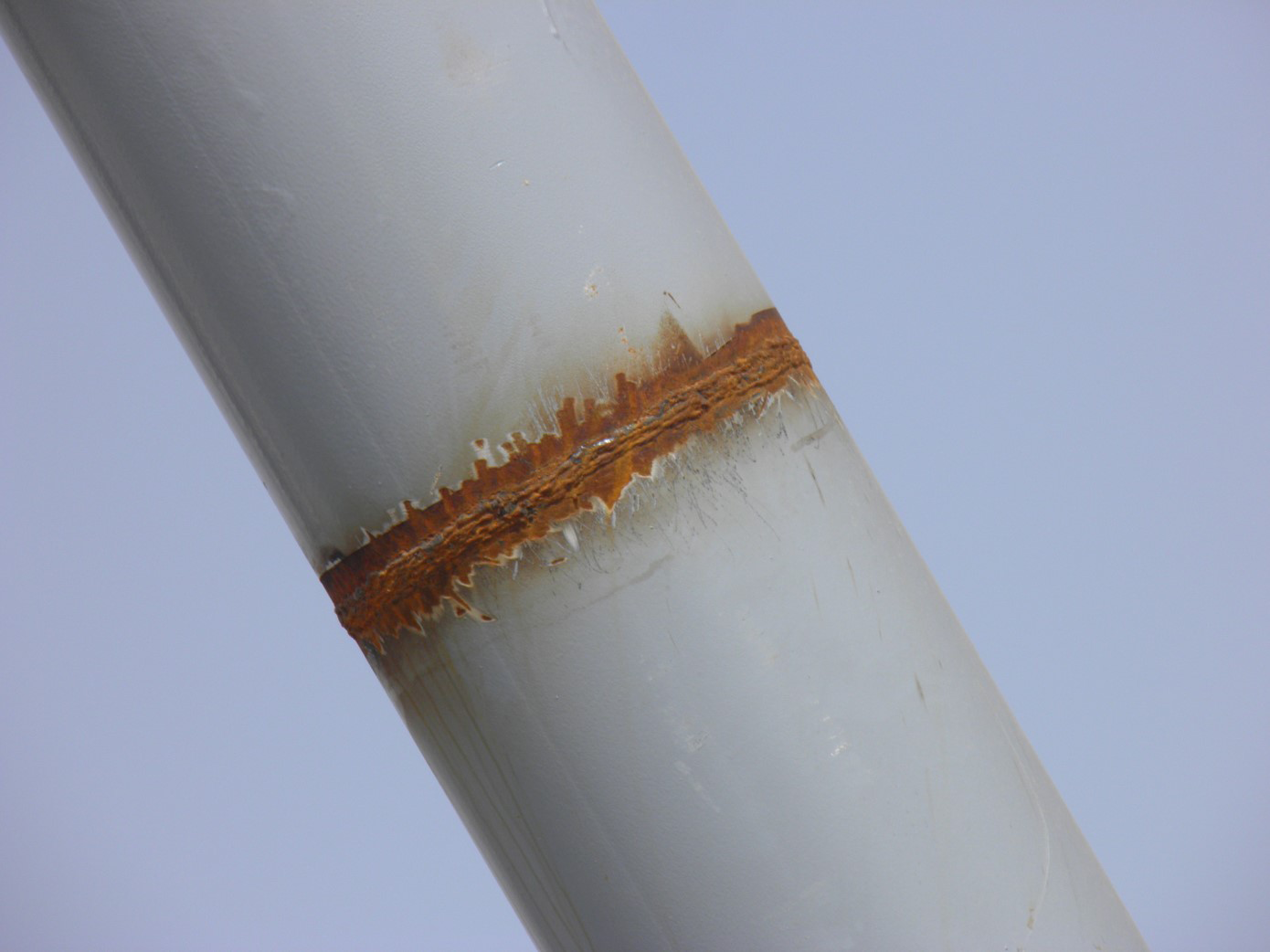}}
  \vspace{0.1cm}
\end{minipage}
\hfill
\begin{minipage}[b]{0.18\linewidth}
  \centering
  \centerline{\includegraphics[width=1.6cm]{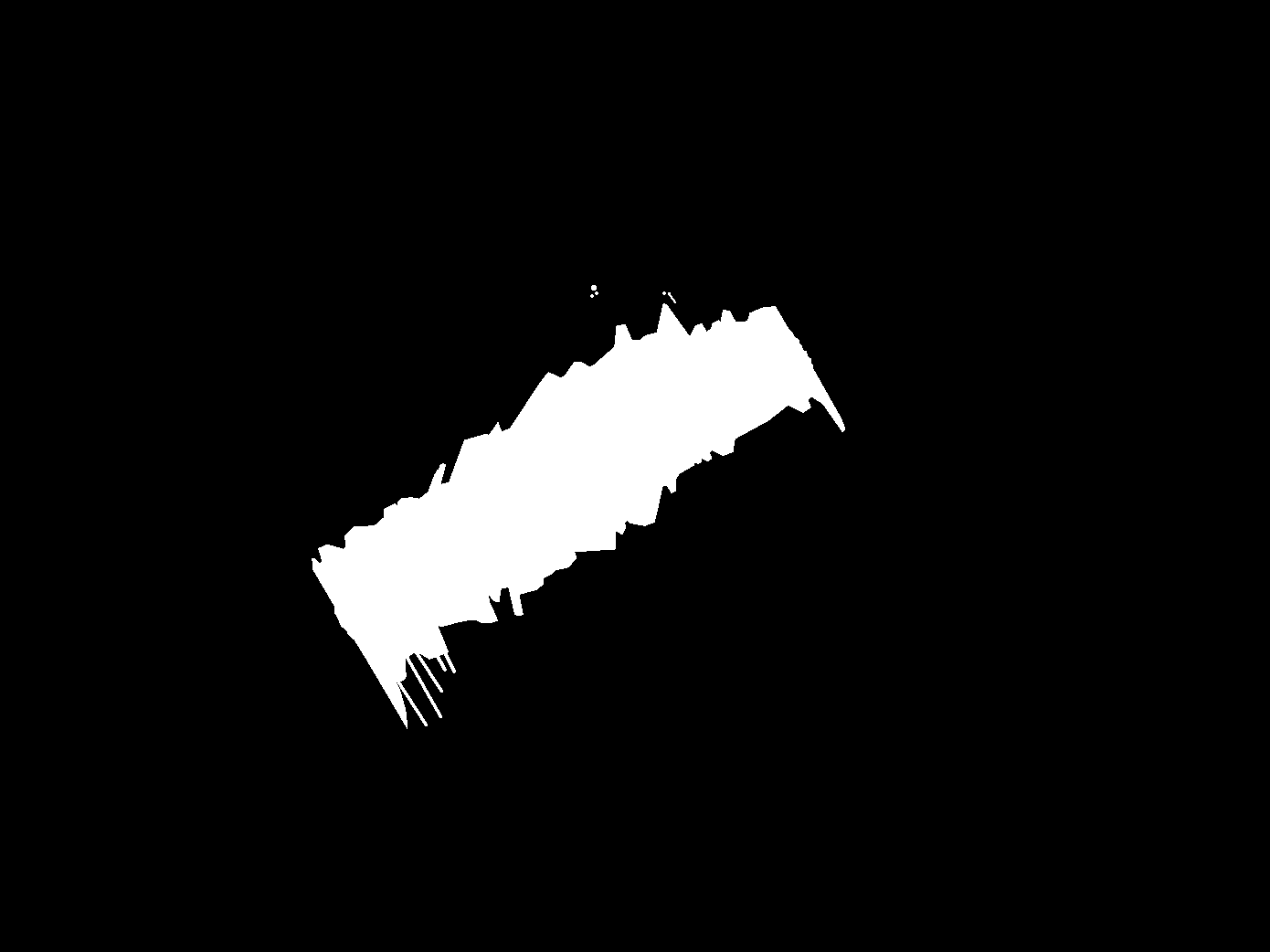}}
  \vspace{0.1cm}
\end{minipage}
\hfill
\begin{minipage}[b]{0.18\linewidth}
  \centering
  \centerline{\includegraphics[width=1.6cm]{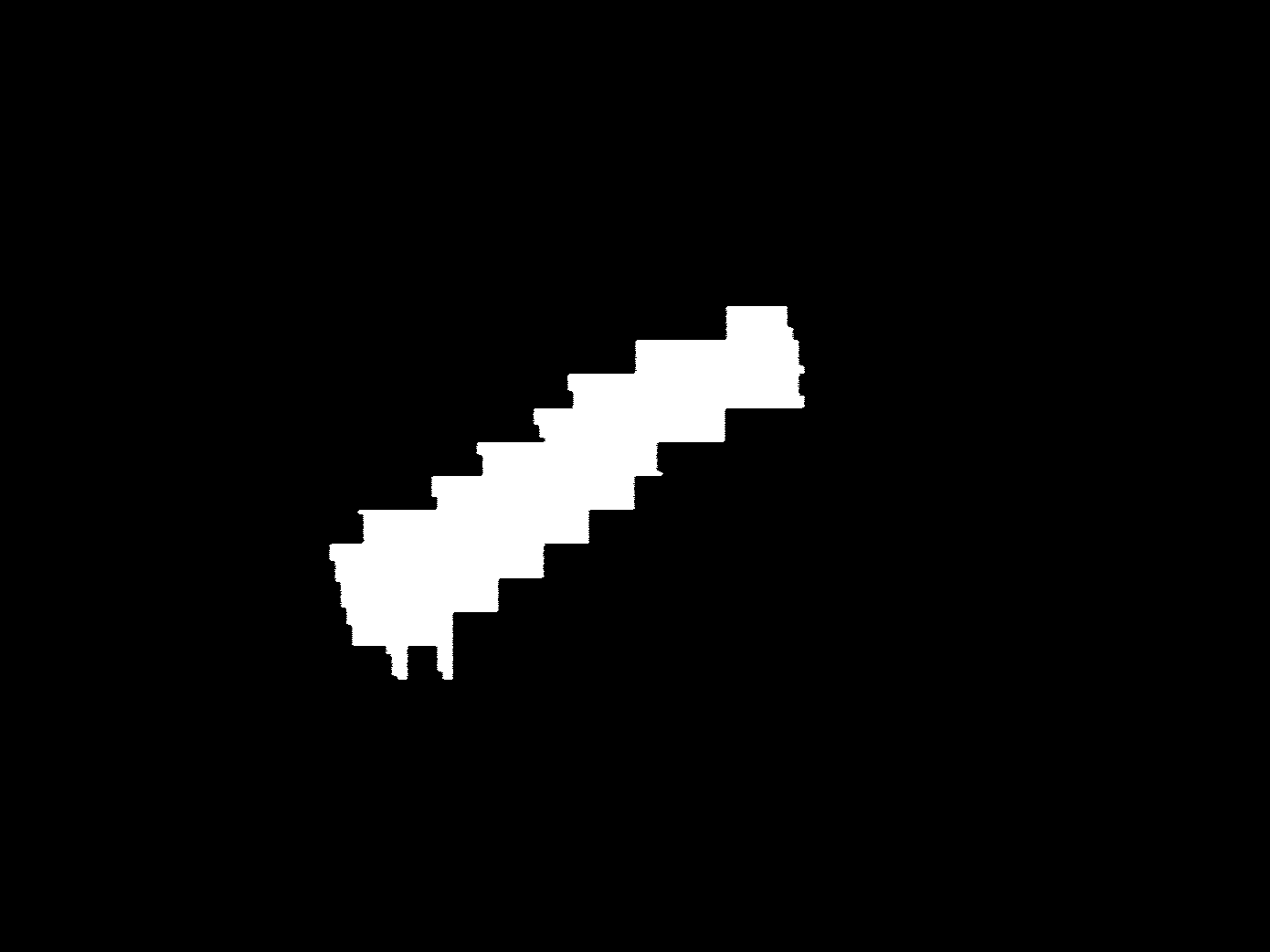}}
  \vspace{0.1cm}
\end{minipage}
\hfill
\begin{minipage}[b]{0.18\linewidth}
  \centering
  \centerline{\includegraphics[width=1.6cm]{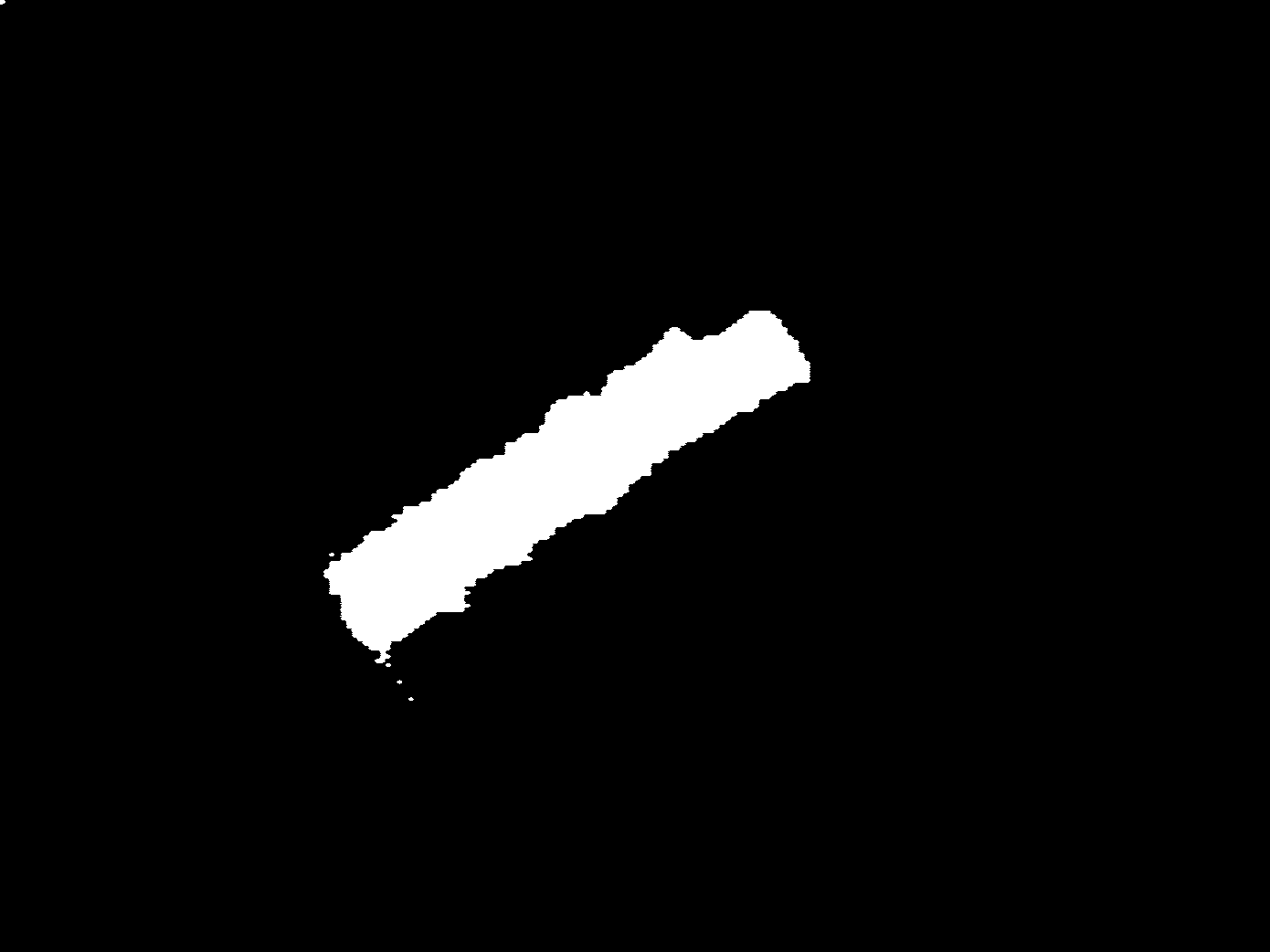}}
  \vspace{0.1cm}
\end{minipage}
\hfill
\begin{minipage}[b]{0.18\linewidth}
  \centering
  \centerline{\includegraphics[width=1.6cm]{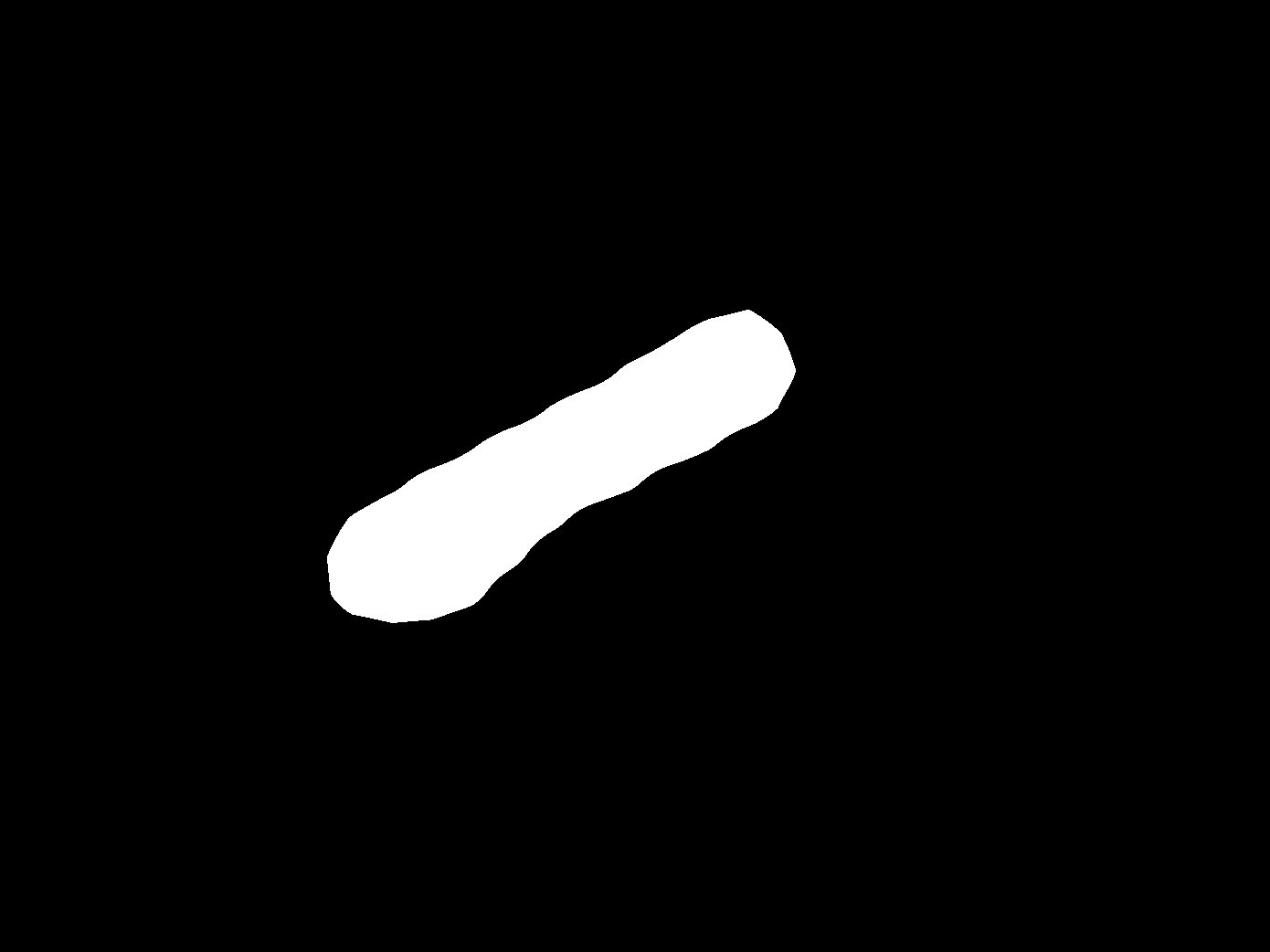}}
  \vspace{0.1cm}
\end{minipage}
\begin{minipage}[b]{0.18\linewidth}
  \centering
  \centerline{\includegraphics[width=1.6cm]{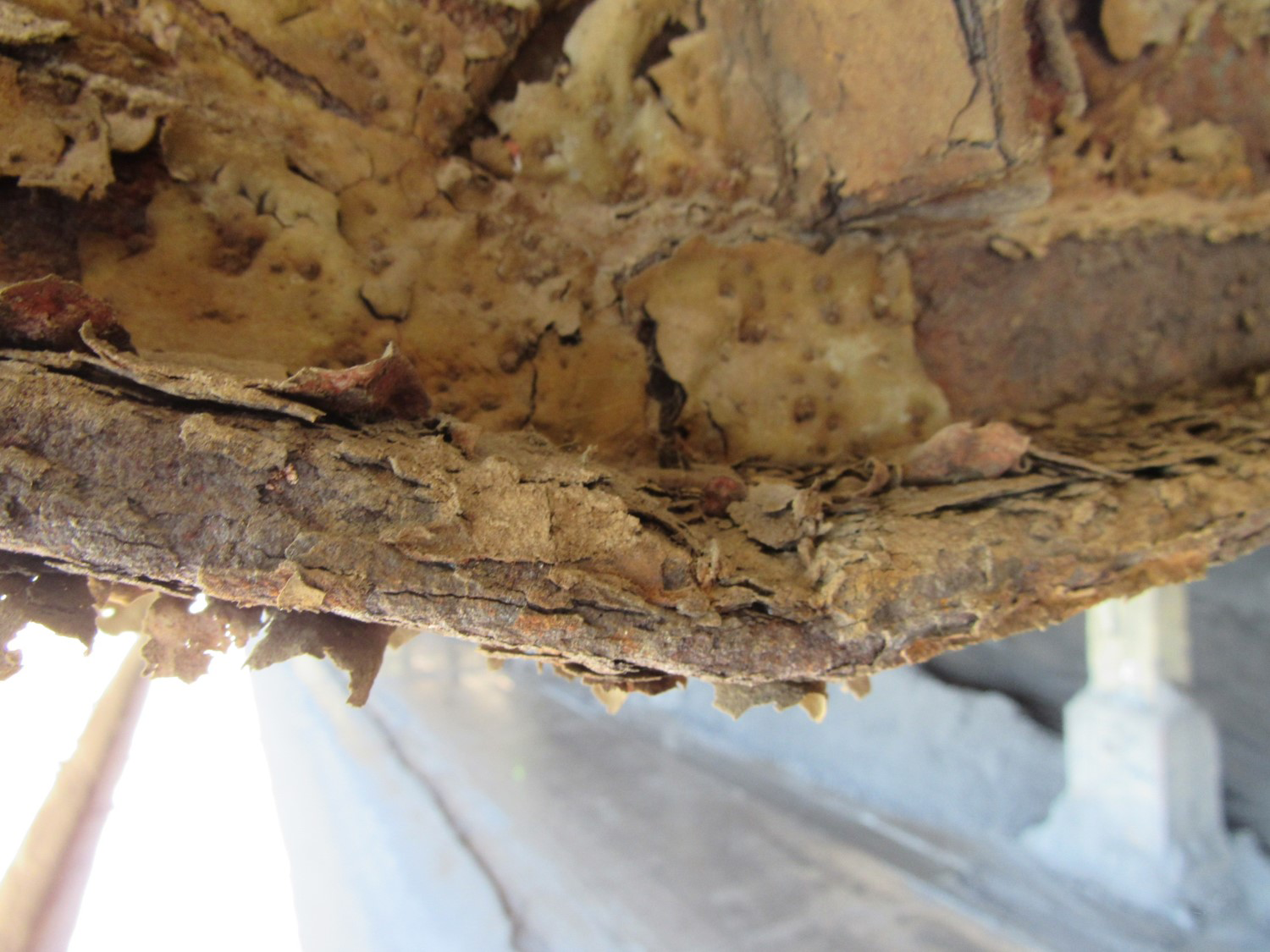}}
  \centerline{Input}\vspace{0.14cm}
\end{minipage}
\hfill
\begin{minipage}[b]{0.18\linewidth}
  \centering
  \centerline{\includegraphics[width=1.6cm]{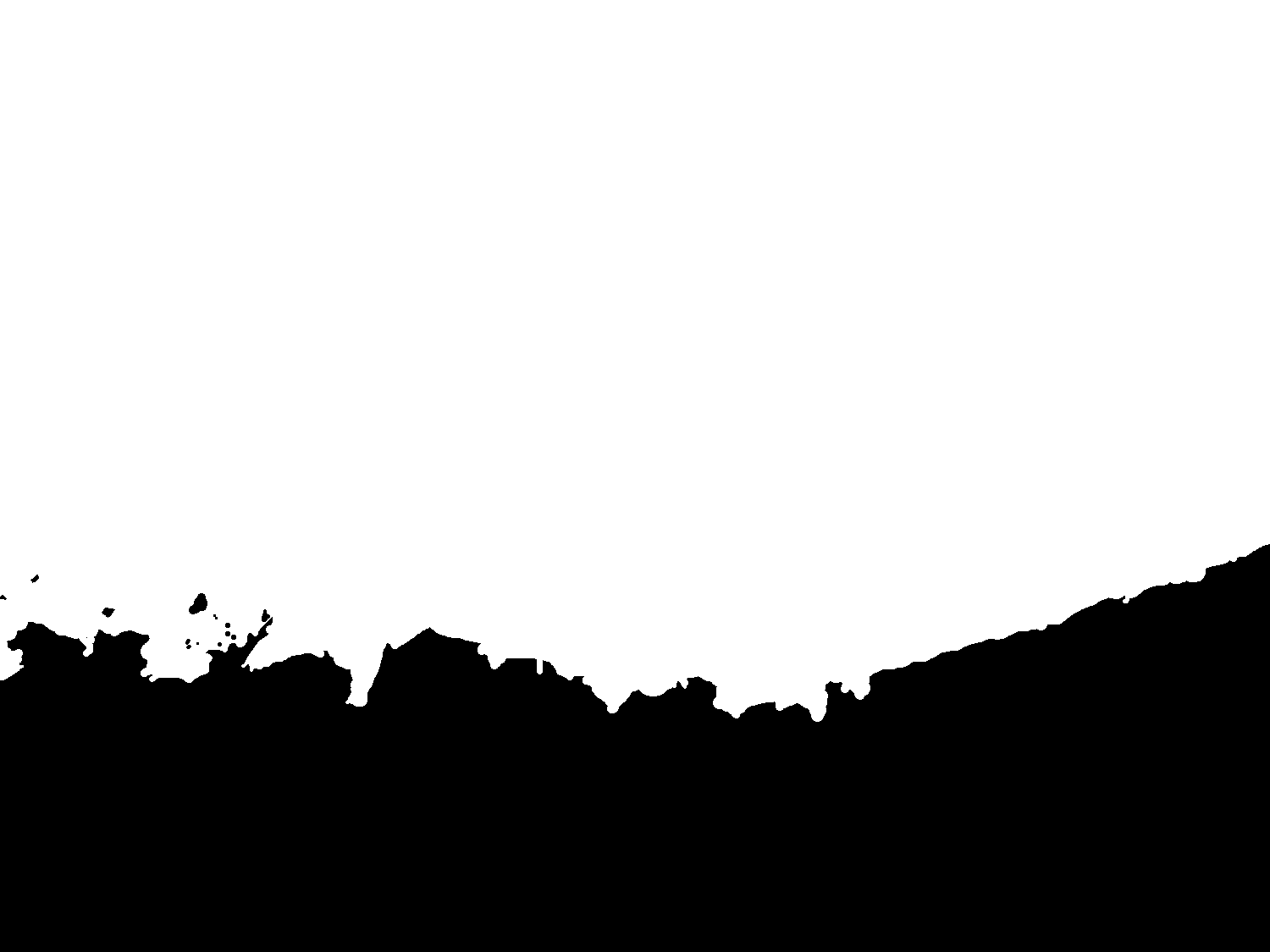}}
 \centerline{Ground truth}\medskip
\end{minipage}
\hfill
\begin{minipage}[b]{0.18\linewidth}
  \centering
  \centerline{\includegraphics[width=1.6cm]{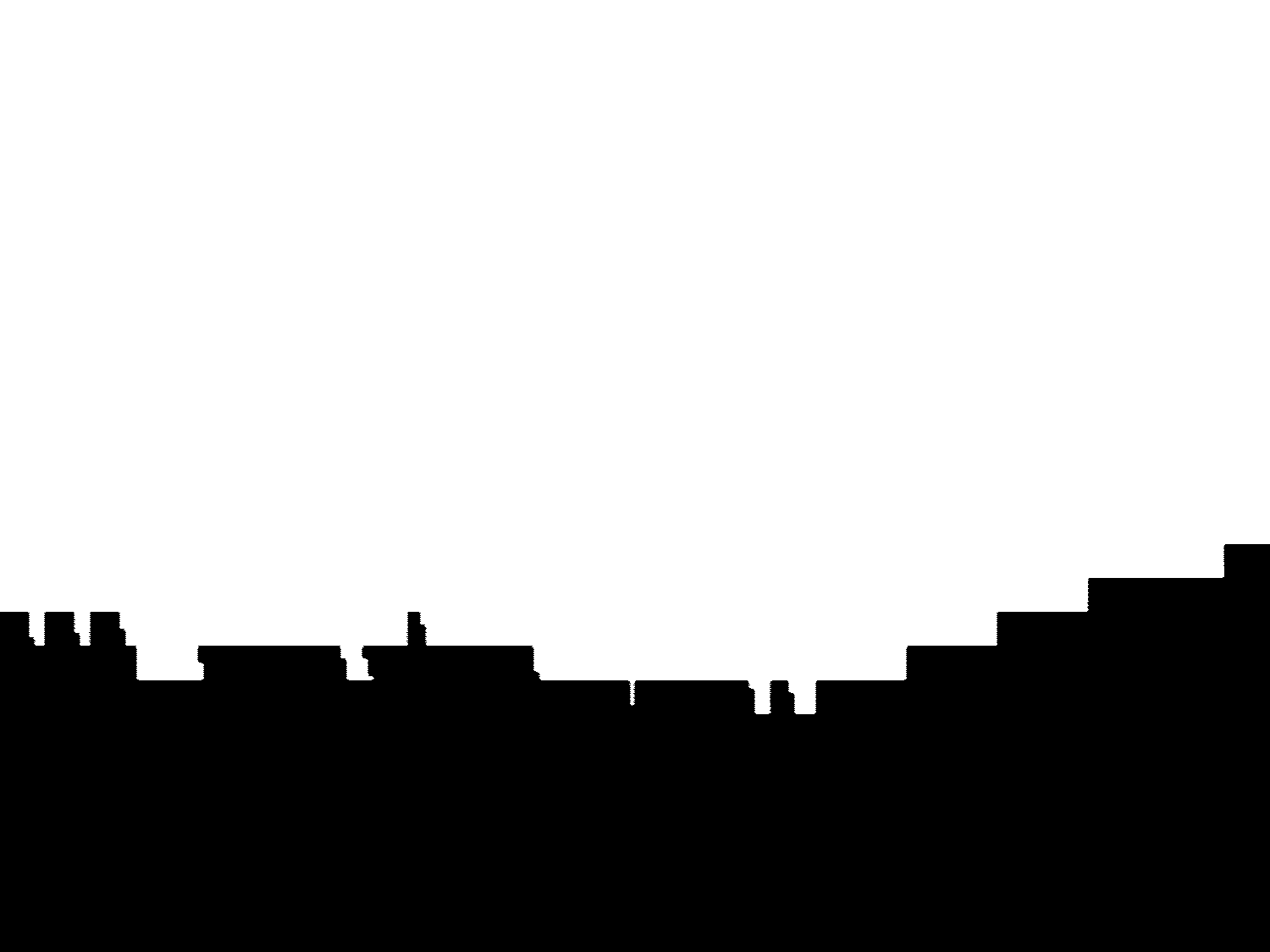}}
  \centerline{FCN}\medskip
\end{minipage}
\hfill
\begin{minipage}[b]{0.18\linewidth}
  \centering
  \centerline{\includegraphics[width=1.6cm]{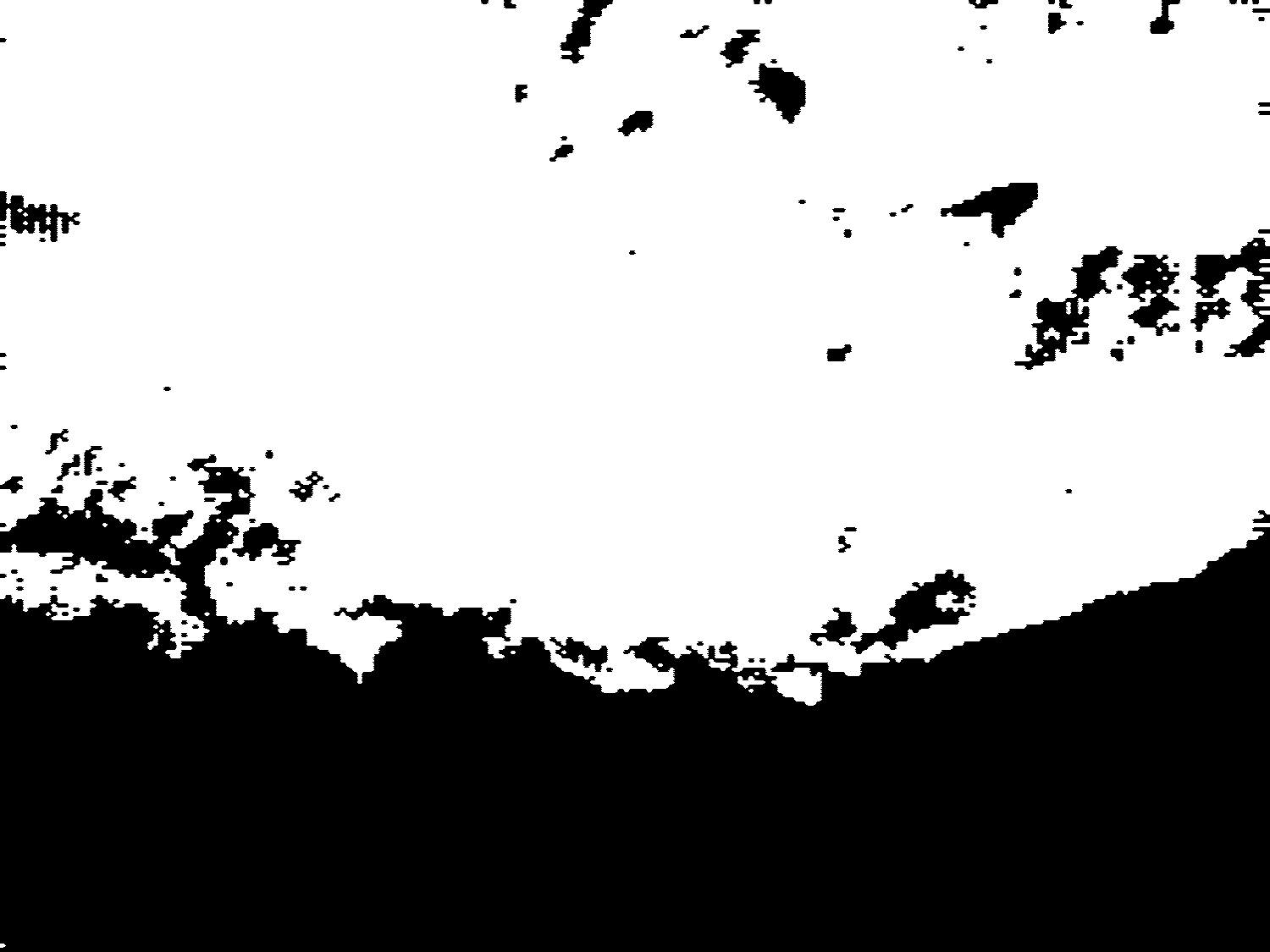}}
  \centerline{U-Net}\medskip
\end{minipage}
\hfill
\begin{minipage}[b]{0.18\linewidth}
  \centering
  \centerline{\includegraphics[width=1.6cm]{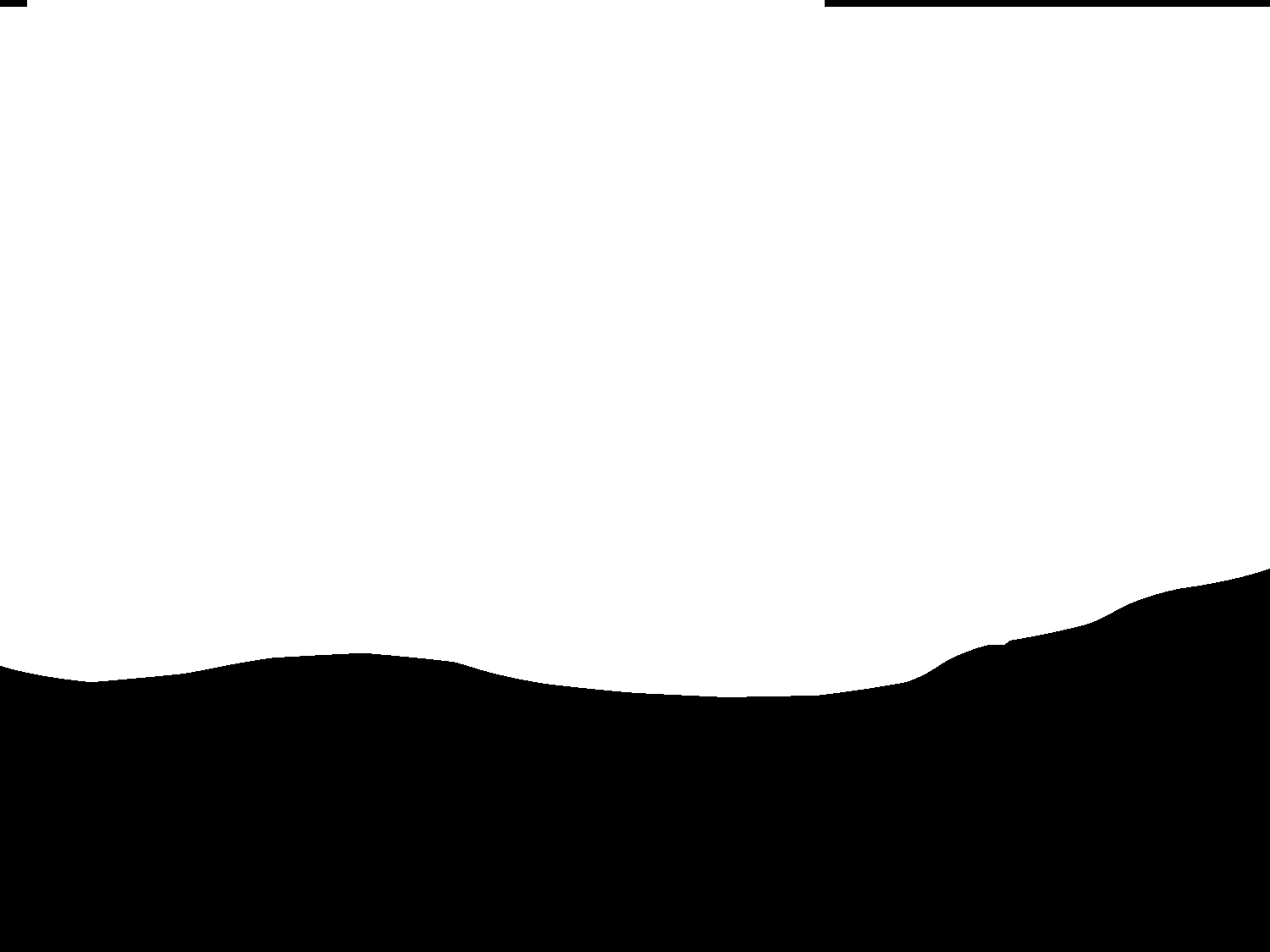}}
  \centerline{Mask R-CNN}\medskip
\end{minipage}
\caption{Visual comparison of the deep models' outputs, without applying boundary refinement.}
\label{fig:deeplearning}
\end{figure}

\subsection{Experimental Results}
\label{sec:expres}

\begin{figure}[h]
\begin{minipage}[b]{0.24\linewidth}
  \centering
  \centerline{\includegraphics[width=1.9cm]{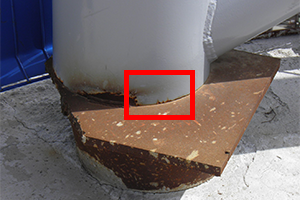}}
  \vspace{0.1cm}
\end{minipage}
\hfill
\begin{minipage}[b]{0.24\linewidth}
  \centering
  \centerline{\includegraphics[width=1.9cm]{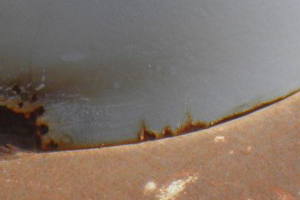}}
  \vspace{0.1cm}
\end{minipage}
\hfill
\begin{minipage}[b]{0.24\linewidth}
  \centering
  \centerline{\includegraphics[width=1.9cm]{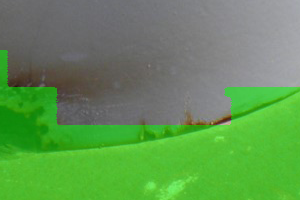}}
  \vspace{0.1cm}
\end{minipage}
\hfill
\begin{minipage}[b]{0.24\linewidth}
  \centering
  \centerline{\includegraphics[width=1.9cm]{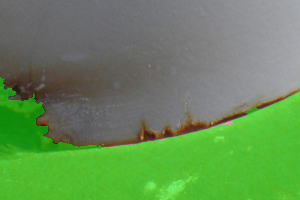}}
  \vspace{0.1cm}
\end{minipage}
\begin{minipage}[b]{0.24\linewidth}
  \centering
  \centerline{\includegraphics[width=1.9cm]{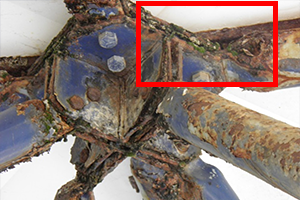}}
  \vspace{0.1cm}
\end{minipage}
\hfill
\begin{minipage}[b]{0.24\linewidth}
  \centering
  \centerline{\includegraphics[width=1.9cm]{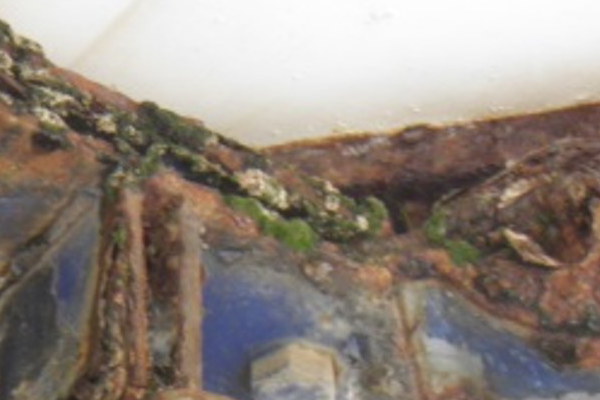}}
  \vspace{0.1cm}
\end{minipage}
\hfill
\begin{minipage}[b]{0.24\linewidth}
  \centering
  \centerline{\includegraphics[width=1.9cm]{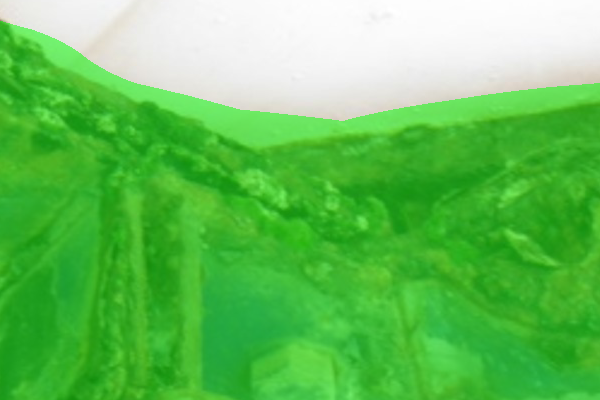}}
  \vspace{0.1cm}
\end{minipage}
\hfill
\begin{minipage}[b]{0.24\linewidth}
  \centering
  \centerline{\includegraphics[width=1.9cm]{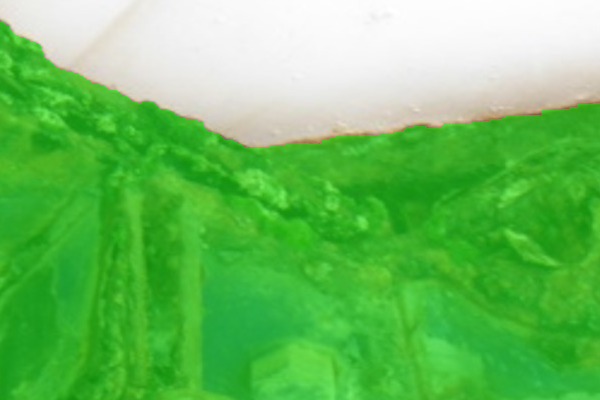}}
  \vspace{0.1cm}
\end{minipage}
\begin{minipage}[b]{0.24\linewidth}
  \centering
  \centerline{\includegraphics[width=1.9cm]{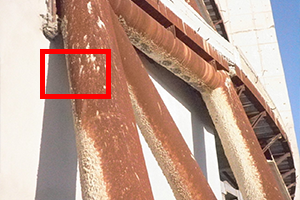}}
  \centerline{Input}\medskip
\end{minipage}
\hfill
\begin{minipage}[b]{0.24\linewidth}
  \centering
  \centerline{\includegraphics[width=1.9cm]{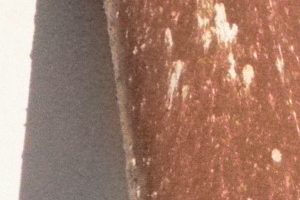}}
  \centerline{Focus area}\vspace{.283cm}
\end{minipage}
\hfill
\begin{minipage}[b]{0.24\linewidth}
  \centering
  \centerline{\includegraphics[width=1.9cm]{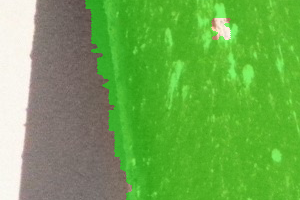}}
  \centerline{Deep output}\medskip
\end{minipage}
\hfill
\begin{minipage}[b]{0.24\linewidth}
  \centering
  \centerline{\includegraphics[width=1.9cm]{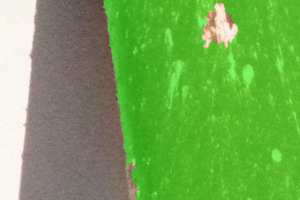}}
  \centerline{Final output}\medskip
\end{minipage}
\caption{Boundary refinement outputs for specific areas.}
\label{fig:overall_model_images}
\end{figure}

Fig. \ref{fig:deeplearning} demonstrates two corroded images along with the respective ground truth annotation for all of the three deep models, without applying boundary refinement. In general, the results depict the semantic segmentation capabilities of the core models. Nevertheless, boundary regions are rather coarse. To refine these areas, we implement the color projection methodology as described in Section \ref{sec:postprocessing} and visualized in Fig. \ref{fig:overall_model_images}. In particular, it shows the defected regions before and after the boundary refinement (see the last two columns of Fig. \ref{fig:overall_model_images}). The corroded regions are illustrated in green for better clarification.  

\begin{figure}[htb!]
  \centerline{\includegraphics[width=8.5cm]{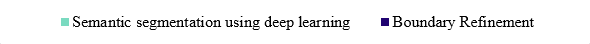}}
  
  \begin{minipage}[b]{0.48\linewidth}
  \centering
  \centerline{\includegraphics[width=3.7cm]{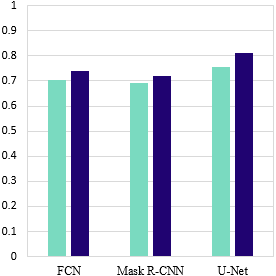}}
  \centerline{(a) Precision}\medskip
\end{minipage}
\hfill
\begin{minipage}[b]{0.48\linewidth}
  \centering
  \centerline{\includegraphics[width=3.7cm]{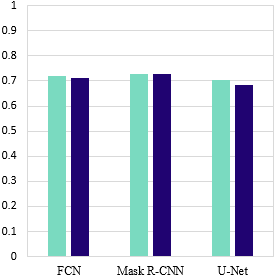}}
  \centerline{(b) F1-Score}\medskip
\end{minipage}
  
  \caption{Comparative performance metrics for the different deep models, before and after the boundary refinement.}
  \label{fig:metrics}
\end{figure}

Objective results are depicted in Fig. ~\ref{fig:metrics} using precision and F1-score. The proposed fusion method improves precision and slightly the F1-score. Mask R-CNN performs slightly better in terms of precision than the rest ones, while for F1-score all deep models yield almost the same performance. Investigating the time complexity, U-Net is the fastest approach, followed by FCN (1.79 times slower) and last is Mask R-CNN (18.25 times slower). Thus, even if Mask R-CNN followed by boundary refinement performs better than the rest in terms of F1-score, it may not be used as the main detection mechanism due to the high execution times.  

\section{Conclusions}
\label{sec:conc}
In this paper, we propose a novel data projection scheme to yield accurate pixel-based detection of corrosion regions on metal constructions. This projection/fusion exploits the results of deep models, which correctly identify the semantic area of a defect but fail on the boundaries, and a color segmentation algorithm which over-segments the defect into multiple color areas but retains contour accuracy. The deep models were FCN, U-Net and Mask R-CNN. Experimental results and comparisons on real datasets verify the out-performance of the proposed scheme, even for very tough image content of multiple types of defects. The performance is evaluated on a dataset annotated by engineer experts. Though the increase in accuracy is relatively small, the new defected areas can significantly improve structural analysis and pre-fabrication than other traditional methods.

\bibliographystyle{IEEEbib}
\bibliography{strings,refs}

\end{document}